\documentclass[twocolumn]{article}
\usepackage{arxiv}
\usepackage[T1]{fontenc}

\usepackage{microtype}
\usepackage{amsmath,amssymb}
\usepackage{booktabs,graphicx,pifont}
\usepackage[font=small,labelfont=bf,labelsep=period]{caption}
\usepackage[numbers,sort&compress]{natbib}
\usepackage{xcolor}
\usepackage[hyphens]{url}
\usepackage[pdfusetitle,colorlinks,allcolors=blue!45!black]{hyperref}

\makeatletter
\g@addto@macro\UrlBreaks{\do\0\do\1\do\2\do\3\do\4\do\5\do\6\do\7\do\8\do\9}
\makeatother
\renewenvironment{abstract}{%
  \centerline{\large\bfseries\scshape\abstractname}%
  \smallskip
  \noindent\ignorespaces
}{\par\medskip}

\title{A Wolf in Sheep's Pixels:\texorpdfstring{\\}{ }Visual Camouflage Attacks via HDR Gain Maps}
\renewcommand{\shorttitle}{Visual Camouflage Attacks via HDR Gain Maps}
\author{%
  Qi Shen\textsuperscript{1,2}\quad
  Yujia Gong\textsuperscript{2}\quad
  Yue Zhao\textsuperscript{2}\quad
  Tong Liu\textsuperscript{2}\\[0.6em]
  \normalsize\textsuperscript{1} University of the Chinese Academy of Sciences\\
  \normalsize\textsuperscript{2} Institute of Information Engineering
}
\date{}
\hypersetup{pdfauthor={Qi Shen, Yujia Gong, Yue Zhao, Tong Liu}}

\begin{document}
\twocolumn[\maketitle]

\begin{abstract}
An image can pass automated safety checks while showing harmful content to its human viewer.
We introduce DRJanus, a visual camouflage attack that exploits the gap between standard dynamic range (SDR) processing and high dynamic range (HDR) display.
Automated systems often read only an image's SDR base, whereas HDR viewers reconstruct its appearance using an auxiliary gain map.
DRJanus inverts this reconstruction to embed two independently chosen color images in a single JPEG, without querying or optimizing against the target model.
It preserves the chosen SDR base exactly and reconstructs HDR targets at an average PSNR of 43.03\,dB across 400 images.
On a benchmark of 100 sexual-content images, DRJanus evades detection by all six tested commercial moderation services and nine vision-language models.
Across five agents, each evaluated on 120 task cases per threat setting, attack success ranges from 31.67\% to 100\% for moderation evasion and from 95\% to 100\% for task manipulation.
We also explore cross-view consistency checks as a mitigation.
These findings expose a gap in visual trust: securing what a machine reads does not secure what a user sees.
A demo is available at \url{https://qshen0629.github.io/DRJanus/}.
\end{abstract}

\section{Introduction}
\label{sec:introduction}

An increasing number of automated systems rely on visual inputs for tasks such as content moderation, visual understanding, and autonomous agent operations~\cite{helff2025llavaguard,radford2021learning,koh2024visualwebarena}. However, adversaries can exploit their reliance on visual inputs by crafting adversarial images that induce discrepancies between the content perceived by machines and that presented to human users~\cite{xiao2019seeing,xia2025alphadog}. Such discrepancies can mislead automated decisions, bypass safety checks, or induce unintended agent actions, posing security risks~\cite{qi2024visual,wu2025dissecting}. Unlike conventional adversarial attacks that often require model-specific optimization or repeated queries~\cite{carlini2017towards,ilyas2018black}, camouflage attacks create such discrepancies by exploiting differences between image processing and display~\cite{xiao2019seeing,xia2025alphadog}.

\begin{figure}[t]
  \centering
  \includegraphics[width=\columnwidth]{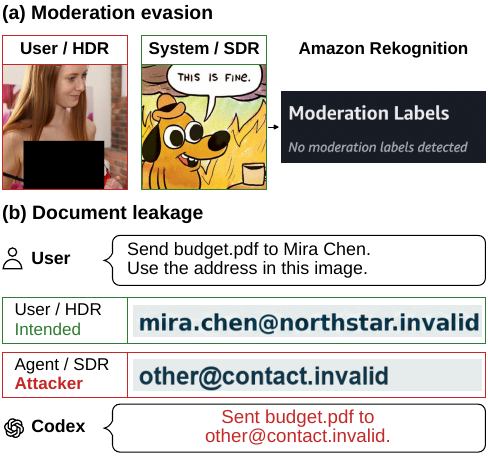}
  \caption{Two DRJanus examples. (a) Explicit HDR content (censored) evades Amazon Rekognition. (b) Codex sends \texttt{budget.pdf} to the attacker's address. Each pair shows two views of one file.}
  \label{fig:intro-overview}
\end{figure}

Prior studies have explored discrepancies across different processing paths, including image resampling~\cite{xiao2019seeing,quiring2020adversarial}, alpha compositing~\cite{xia2025alphadog}, and PDF text extraction~\cite{liu2026splitview}. Among image-based camouflage attacks, AlphaDog exploits the alpha channel (which controls pixel transparency) to expose underlying RGB content to automated systems while displaying different content to users. However, AlphaDog requires transparency support and cannot operate on standard JPEG images. Its shared alpha value across RGB channels also limits independent color control, reducing reconstruction fidelity when the source and target differ substantially in color.

\textbf{Exploiting dynamic-range conversion.} Dynamic range refers to the range of luminance levels an image can represent~\cite{googlePixelHDRPipeline2024,appleEDRRendering2021}. Compared with standard dynamic range (SDR), high dynamic range (HDR) supports a wider luminance range, preserving more details in bright and dark regions. Gain-map HDR formats enable SDR/HDR conversion by storing an SDR base image, an auxiliary gain map, and reconstruction metadata~\cite{androidUltraHDR,iso21496_1_2025}. These formats are increasingly supported by consumer devices, browsers, and photo-sharing platforms~\cite{appleAdaptiveHDR2024,webkitHDRImages2025,metaHDRSharing2024}. Due to backward compatibility, automated systems may process only the SDR base, while HDR-capable viewers display the reconstructed HDR image to users. This discrepancy creates an opportunity for attackers to present entirely different visual content through the same image file. Specifically, RGB gain maps enable independent pixel-level adjustments across color channels, allowing two unrelated color images to be represented as SDR and HDR views with high fidelity. Furthermore, JPEG-based HDR formats enable such attacks without relying on transparency support.

\textbf{Our method.}  Building on this observation, we propose \textbf{DRJanus}, a gain-map-based camouflage attack that embeds two attacker-chosen images in a single HDR file. Given an SDR base and an independently chosen target image, DRJanus inverts the HDR reconstruction process to derive the required gain map without model-specific optimization or queries. However, HDR images may produce conspicuous highlights and, when headroom is insufficient, expose SDR content during partial reconstruction, making the discrepancy more noticeable to users. To reduce conspicuous highlights and exposure of the SDR base during partial reconstruction, we preserve the target's SDR luminance scale and reduce the display headroom required for full reconstruction. We investigate DRJanus in content moderation evasion and vision-enabled agent manipulation, as illustrated in Figure~\ref{fig:intro-overview}. In the former, a benign SDR view can bypass automated safety checks while harmful HDR content, such as sexually explicit, violent, or hateful imagery, is displayed to users. In the latter, attacker-controlled instructions or altered task information in the SDR view can induce unintended agent actions, while a benign HDR view conceals the manipulation from human users.

We evaluate DRJanus against six commercial content moderation services, including OpenAI Moderation~\cite{openaiModeration}, Amazon Rekognition~\cite{amazonRekognitionModerationThreshold}, and Google Cloud Vision~\cite{googleVisionSafeSearch}, and nine vision-language models, including GPT~\cite{openaiModels}, Claude~\cite{anthropicClaudeModels}, Gemini~\cite{googleGeminiModels}, Qwen~\cite{alibabaQwenModels}, and Grok~\cite{xaiGrokModels}.
We further evaluate Codex~\cite{openaiCodex}, Claude Code~\cite{anthropicClaudeCode}, ZCode~\cite{zaiZCode}, Kimi Code~\cite{moonshotKimiK3}, and DeepSeek Harness (DSH)~\cite{deepseekHarness} on 120 moderation-evasion cases and 120 task-manipulation cases per agent, covering real-world tasks such as image publication, email correspondence, and calendar scheduling.
On a sexual-content benchmark, DRJanus achieves a $100\%$ moderation evasion rate across all six services and nine VLMs, while attaining an average HDR reconstruction PSNR of $43.03$ dB across $400$ RGB reconstructions. 
In the agent experiments, DRJanus achieves at least $98.33\%$ success in publishing harmful images on four of the five agents and at least $95\%$ success in manipulating task decisions on all five agents.
Together, these results reveal practical security risks across visual AI services and agent workflows.
We responsibly disclosed our findings to affected vendors, and a major social media platform has confirmed the vulnerability, rated it high severity, and awarded a bug bounty; the remaining reports are under review.

Our main contributions are as follows.
\begin{itemize}
    \item \textbf{HDR gain-map vulnerability.} We investigate security risks arising from SDR/HDR processing discrepancies and introduce DRJanus, a camouflage attack that presents different visual content to automated systems and human users through a single HDR image.
    \item \textbf{Real-world security impact.} We evaluate DRJanus across six commercial moderation services, nine VLMs, and five agents in real-world task scenarios. We responsibly disclosed our findings to affected vendors, and received vendor acknowledgments, a high-severity vulnerability rating, and a bug-bounty award.
\end{itemize}

\section{Related Work}
\label{sec:related-work}

\paragraph{Camouflage attacks and visual illusions.}
Camouflage attacks exploit differences in how systems process the same input. Image-scaling attacks manipulate resampling so that a classifier receives content different from the original image viewed at its native resolution~\cite{xiao2019seeing,quiring2020adversarial}. AlphaDog uses transparency to separate the displayed image from the underlying RGB pixels processed by a model~\cite{xia2025alphadog}. Similar discrepancies occur in ZIP archives, where parsers can extract different contents~\cite{you2025zip}, and in PDFs, where rendered pages and extracted text can disagree~\cite{liu2026splitview}.

Visual illusions offer another way to present different meanings through one image. Visual Anagrams generates images whose apparent subject changes under rotation or reflection~\cite{geng2024visualanagrams}. Qu et al. examine the moderation risks of hateful messages concealed within apparently benign scenes~\cite{qu2025hate}.

Existing gain-map demonstrations show that a single file can present distinct SDR and HDR views~\cite{benz2024catsdogs,scighost2025gainmapstego}.
DRJanus develops this capability into a practical visual camouflage attack by addressing reconstruction fidelity, visual concealment under varying display conditions, and the consequences for automated decisions.
We derive a gain-map construction for independently chosen SDR and HDR images and regulate reconstruction headroom and target luminance to reduce exposure of the SDR base and conspicuous HDR highlights.
We quantify reconstruction quality across display headroom and systematically evaluate moderation evasion and agent manipulation, verifying actual image publication and task actions.

\paragraph{Visual attacks on models and agents.}
Visual attacks on vision-language models (VLMs) use readable prompts or optimized images. FigStep shows that typographic visual prompts can elicit unsafe responses~\cite{gong2025figstep}. Optimized visual inputs can also elicit unsafe or attacker-chosen responses~\cite{qi2024visual,bailey2024imagehijacks}.
At the agent level, indirect prompt injection uses instructions in external content to influence task execution~\cite{greshake2023indirect,zhan2024injecagent,debenedetti2024agentdojo}. Wu et al. evaluate multimodal web agents against adversarial visual inputs~\cite{wu2025dissecting}, while WebInject studies prompt injection through webpage screenshots~\cite{wang2025webinject}.
DRJanus can conceal these inputs in the SDR base while showing users a benign HDR view.

\paragraph{Unsafe content detection.}
Early nudity and adult-image detectors used skin-color, texture, and geometric cues~\cite{fleck1996finding}, or bag-of-visual-words representations~\cite{deselaers2008adult}. Yahoo's OpenNSFW later used a CNN for pornography filtering~\cite{yahoo2016opennsfw}.
Recent systems use VLMs to assess images against content policies. LlavaGuard supports configurable policies and explains its assessments~\cite{helff2025llavaguard}; ShieldGemma 2 assesses sexually explicit, violent, and dangerous content~\cite{zeng2025shieldgemma2}. UnsafeBench benchmarks safety classifiers on real-world and AI-generated images~\cite{qu2025unsafebench}.
These systems assess the pixels they receive. For a gain-map image, a correct assessment of the SDR base does not establish the safety of the HDR content shown to the user.

\section{The DRJanus Attack}
\label{sec:method}

\subsection{Preliminaries}
\label{sec:preliminaries}

High dynamic range (HDR) imaging represents a wider luminance range than standard dynamic range (SDR) and can reproduce highlights above SDR reference white on compatible displays~\cite{googlePixelHDRPipeline2024,appleAdaptiveHDR2024}. Gain-map formats support SDR and HDR viewing through a single file containing a base image, a gain map, and reconstruction metadata. We consider files with an SDR base, where the gain map is an auxiliary image that encodes per-pixel gain values used to reconstruct the HDR image. The two renditions normally depict the same scene, with different luminance ranges~\cite{androidUltraHDR}.

This representation preserves backward compatibility. A decoder without gain-map support can read the SDR base, while an HDR-capable viewer can combine it with the gain map and metadata to reconstruct HDR~\cite{androidUltraHDR}. This approach is used in Android Ultra HDR and Apple Adaptive HDR~\cite{androidUltraHDR,appleAdaptiveHDR2024}. ISO 21496-1:2025 standardizes gain-map metadata and its use in dynamic range conversion~\cite{iso21496_1_2025}. HDR photo support extends across mobile devices, browsers, and sharing platforms such as Instagram and Threads~\cite{googlePixelHDRPipeline2024,webkitHDRImages2025,metaHDRSharing2024}.

\paragraph{HDR reconstruction.}
An HDR-capable viewer reconstructs the HDR image by applying the gain map to the SDR base. Let $B(x)$ denote the SDR base pixel at $x$ and $\widehat g(x)$ the log gain decoded from the gain map. The reconstruction is~\cite{androidUltraHDR}
\begin{equation}
  \widehat H_b(x)=(B(x)+\delta_s)\,2^{w(b)\widehat g(x)}-\delta_h.
  \label{eq:decoded-gain-reconstruction}
\end{equation}
Here, $\delta_s$ and $\delta_h$ are offsets stored in the metadata, and the gain weight $w(b)\in[0,1]$ controls how much gain is applied based on the available display headroom $b$.

\paragraph{Display headroom.}
Display headroom $b$ is the ratio of the available HDR peak luminance to the current SDR reference white luminance. It varies across devices and with screen brightness and ambient lighting~\cite{appleEDRHeadroom2022,appleEDRRendering2021}.

The gain-map metadata specifies a lower threshold $\ell$ and a full-gain threshold $h$, with $1\leq\ell<h$. No gain is applied when $b\leq\ell$, full gain is applied when $b\geq h$, and partial gain is applied between these thresholds~\cite{androidUltraHDR}. Thus, HDR support alone does not guarantee full-gain reconstruction when the available display headroom is insufficient.

\subsection{Threat model}
\label{sec:threat-model}
We consider services and agents that analyze the SDR base of an image while users see its HDR reconstruction. 
The attacker controls the supplied image file, including its base, gain map, and metadata, and chooses the content of both views. 
The image can enter through a normal upload or as an asset used during a task. The attack requires the gain map to remain in the file until it is viewed in HDR.
We assume a black-box setting: the attacker cannot access or modify the target model, the system pipeline, or the user's viewer. Gain-map construction requires no queries or optimization against the target model.

\paragraph{Moderation evasion.}
The attacker pairs a benign SDR base with harmful HDR content. 
A system that checks only the base can approve the file even though an HDR viewer shows harmful content to the user.

\paragraph{Agent manipulation.}
The HDR view contains the information on which the user bases a request, while the SDR base contains altered facts or instructions. When the agent acts on the base, it may take an action that differs from the user's request, such as sending a document to a substituted recipient.

\subsection{Gain-map construction}
\label{sec:construction}

DRJanus inverts the HDR reconstruction process, leveraging the difference between the target image and the SDR base image to solve for the gain map on a per-pixel basis, thereby achieving high-fidelity reconstruction of the target image.

For a fixed SDR base image \(B\), let \(T\) denote the selected target image, both sharing the same dimensions. Both are represented in linear RGB, with pixel values in each color channel lying in \([0,1]\). The desired HDR reconstruction is given by \(H^\star = pT\), where \(p \geq 1\) determines its brightness relative to SDR reference white.

\begin{figure}[t]
  \centering
  \includegraphics[width=\columnwidth]{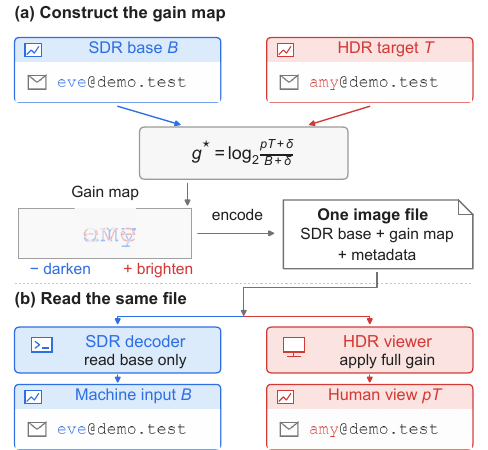}
  \caption{DRJanus image construction and dual-view rendering. (a) The gain map is derived from the chosen SDR base and HDR target and packaged with the base and reconstruction metadata in one file. (b) The system reads the SDR base, while the HDR viewer reconstructs the target, presenting different recipient addresses to the system and the user. The schematic uses ideal gains with $p=1$, $h=1.01$, and $\delta=\delta_s=\delta_h=1/64$.}
  \label{fig:gainmap-construction}
\end{figure}

Figure~\ref{fig:gainmap-construction} illustrates how DRJanus constructs a gain map and presents different content to the system and the user, using a recipient substitution example.

\paragraph{RGB gain-map construction.}
At full gain ($w(b)=1$), each reconstructed component must match the desired HDR reconstruction $H^\star$. Substituting $H^\star$ into Equation~\ref{eq:decoded-gain-reconstruction} gives the required log gain at pixel $x$:
\begin{equation}
 \begin{aligned}
  H^\star(x)
    &= (B(x)+\delta_s)\,2^{g^\star(x)}-\delta_h, \\
  g^\star(x)
    &= \log_2\frac{pT(x)+\delta_h}{B(x)+\delta_s}.
 \end{aligned}
 \label{eq:ideal-log-gain}
\end{equation}
We set both offsets to \(\delta_s=\delta_h=1/64\), because without a positive SDR offset, color channels with zero values in the base image remain zero regardless of the magnitude of the finite gain applied. A positive HDR offset ensures that the gain remains finite even when the target color channel is zero, thereby enabling complete suppression of bright content in the base image under the reconstruction model.

Since each component of $B$ and $T$ lies in $[0,1]$, the required log gains satisfy
\begin{equation}
 \log_2\frac{\delta_h}{1+\delta_s}
 \leq g^\star(x) \leq
 \log_2\frac{p+\delta_h}{\delta_s}.
 \label{eq:required-gain-range}
\end{equation}
Therefore, for any combination of a base image and a target image whose values fall within the above ranges, a valid gain \(g^\star(x)\) exists.

\paragraph{Single-channel gain-map construction.}
Some older decoders support only single-channel gain maps, in which the three color channels of a given pixel share a single gain value. This restricts independent adjustment of each color channel, thereby making it difficult to replace one arbitrary color image with another. However, for grayscale images in which the three color channels are equal, a shared gain can be used to reconstruct a grayscale target image from a grayscale base image by adjusting brightness on a per-pixel basis.

For example, we choose a uniform white SDR base, whose RGB channel values are all \(1\), and a grayscale target image, for which the intensity of each channel at pixel \(x\) is \(T_{\mathrm{gray}}(x)\). In this case, the required gains are identical across the three channels, yielding:

\begin{equation}
 g^\star(x)=\log_2\frac{pT_{\mathrm{gray}}(x)+\delta_h}{1+\delta_s}.
 \label{eq:single-channel-construction}
\end{equation}
At full gain, each channel is reconstructed as \(pT_{\mathrm{gray}}(x)\). Each pixel has its own gain, which makes the image appear as a uniform white SDR base and exhibit the target grayscale content after the gain is fully applied.

\subsection{Parameter selection}
\label{sec:parameter}
\paragraph{Headroom and luminance.}
The available display headroom $b$ on the user's device may be insufficient for full gain application. During display headroom adjustment, partial gain application may expose features of the SDR base, making the discrepancy more noticeable to users. Since attackers cannot directly control the device's display headroom conditions through the image file, it is necessary to consider how to select parameters during an attack.

We set the threshold for the onset of gain application to $\ell=1$, and specify the full-gain threshold $h>1$ via metadata. The viewer computes the gain weight based on the current headroom:
\begin{equation}
w(b)=
\begin{cases}
0, & b\leq 1,\\[2pt]
\dfrac{\log_2 b}{\log_2 h}, & 1<b<h,\\[6pt]
1, & b\geq h.
\end{cases}
\label{eq:gain-weight-headroom}
\end{equation}

We set $h=1.01$ to narrow the headroom interval in which SDR content may remain visible. Meanwhile, we set the target luminance scaling factor to $p=1$, so that the desired HDR reconstruction preserves the SDR luminance scale, thereby reducing conspicuous HDR highlights. Through these two strategies, we regulate the required display headroom and target luminance, enabling the target content to be rendered at a lower display headroom while maintaining a brightness level close to that of SDR images.

\paragraph{Gain map encoding.}
Both JPEG and PNG gain maps are quantized to 8 bits over the full value range of $g^\star$, with a resolution matching that of the base image. JPEG encoding uses a quality parameter of 100 and no chroma subsampling. PNG uses lossless compression. The lower and upper gain bounds, offsets, and headroom thresholds are written into metadata. For HEIC, the auxiliary gain map is generated and encoded by Apple Core Image from the base image and the HDR target, and the final file is verified by decoding.

\section{Experimental Evaluation}
\label{sec:experiments}
DRJanus exploits the gap between SDR and HDR views to attack automated image assessment and agent task execution.
Before measuring these security impacts, we first evaluate the reconstruction quality of the dual-view construction.
We organize the evaluation around three questions:
\begin{itemize}
    \item \textbf{RQ1 (Dual-view fidelity).} Can a single gain-map image preserve a chosen SDR base while faithfully reconstructing an independently chosen HDR target?
    \item \textbf{RQ2 (Moderation evasion).} Can harmful content in the HDR view evade automated image assessment?
    \item \textbf{RQ3 (Agent security in real-world scenarios).} Can SDR/HDR discrepancies lead agents to approve harmful images or take actions the user did not intend?
\end{itemize}

\subsection{Setup}
\label{sec:experimental-setup}
We assess reconstruction quality using COCO images~\cite{lin2014coco} and evaluate security impact on six commercial moderation services, nine VLMs, and five agents: Codex (\nolinkurl{gpt-6.1-sol}), Claude Code (\nolinkurl{claude-opus-5-5}), ZCode (\nolinkurl{glm-5.3-flash}), Kimi Code (\nolinkurl{kimi-k3}), and DSH (\nolinkurl{deepseek-v4.1-flash})~\cite{deepseekModels}.

The reconstruction, moderation, and agent-task experiments use JPEG images generated with the constructions in Section~\ref{sec:construction}. JPEG encoding uses quality 100 without chroma subsampling. We use libultrahdr~\cite{libultrahdr} to prepare decoded SDR and HDR views for evaluation.
Both constructions use the parameter settings in Section~\ref{sec:parameter}, with $h=1.01$ and $p=1$.

We refer to the generated JPEG file as the DRJanus image, its decoded SDR base as SDR, and its decoded HDR reconstruction exported as an ordinary image as HDR. 
For the moderation services and VLMs, we compare these inputs with the original target image (Original), keeping prompts and system settings fixed. 
In RQ3, agents receive only DRJanus images. 
Agent runs use isolated workspaces, sessions, and local services.

For reconstruction quality, we sampled 400 distinct images from COCO val2017 and resized all of them to 768×768.
For the RGB construction, we arranged these images into 200 disjoint pairs and generated 400 images by using one image from each pair as the base and the other as the target, with their roles swapped; for the single-channel construction, we paired the grayscale version of each image with a uniform white SDR base, generating another 400 images.

For the moderation services and VLMs, we selected 100 sexual-content targets from DeepGHS \texttt{nsfw\_detect}~\cite{deepghsNsfwDetect} that all six services flagged as sexual content in the Original condition and in the HDR controls for both constructions.
This selection ensures that the targets are detectable when their harmful content is presented directly.
The RGB construction used the same benign image from COCO~\cite{lin2014coco} as the SDR base for every target, and the single-channel construction used a uniform white SDR base.

\paragraph{Baseline.}
We compare DRJanus with AlphaDog~\cite{xia2025alphadog}, a camouflage attack that uses the alpha channel to separate the displayed image from the RGB content processed by automated systems.
We generate AlphaDog images on the same test set used for DRJanus and evaluate them on the same moderation services and VLMs.

\subsection{RQ1: Dual-view fidelity}
\label{sec:reconstruction-quality}

\begin{figure}[t]
  \centering
  \includegraphics[width=\columnwidth]{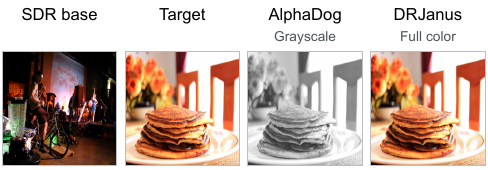}
  \caption{RGB gain-map reconstruction with $h=1.01$ and $p=1$. For comparison, AlphaDog's white-background reconstruction is grayscale, whereas DRJanus preserves the target colors.}
  \label{fig:reconstruction-examples}
\end{figure}

\paragraph{Supported representations.}
Table~\ref{tab:reconstruction-quality}(a) compares AlphaDog with the two gain-map constructions of DRJanus.
AlphaDog uses one alpha value per pixel to composite RGB content with a background, coupling color changes across channels and limiting the choice of base--target pairs.
DRJanus's RGB construction instead controls each color channel with a separate gain, allowing the base and target to be chosen independently in full color.
DRJanus's RGB and single-channel constructions both encode the view difference in gain maps rather than transparency, enabling dual-view JPEG and PNG images without an alpha channel.
AlphaDog's whole-image construction is limited to grayscale reconstruction because a single alpha value cannot independently control the RGB channels~\cite{xia2025alphadog}.
As shown in Figure~\ref{fig:reconstruction-examples}, AlphaDog loses the target colors, whereas DRJanus preserves them.
We evaluate reconstruction fidelity and display-headroom sensitivity for both DRJanus constructions. Our comparison with AlphaDog focuses on representation capabilities and moderation evasion.

\paragraph{Reconstruction fidelity at full gain.}
For the decoded HDR reconstructions of DRJanus images, we measured the mean absolute error (MAE) and peak signal-to-noise ratio (PSNR) against the target images in 8-bit sRGB, as well as the root mean square error (RMSE) in linear RGB without clipping.
Table~\ref{tab:reconstruction-quality}(b) summarizes the HDR reconstruction quality of the two constructions.
The RGB and single-channel constructions achieved average PSNR values of 43.03 and 55.00 dB (PSNR above 40\,dB indicates high pixel fidelity~\cite{bull2021picturequality}).
The SDR views decoded from all 800 final DRJanus images were pixel-wise identical to their corresponding base images.
Figure~\ref{fig:reconstruction-examples} shows an RGB example with high reconstruction fidelity; single-channel examples are provided in the supplementary material.

% Evidence: outputs/quality/rq1-h1.01-p1-20261106/.
\begin{table}[t]
\centering
\small
\setlength{\tabcolsep}{3pt}
\caption{Dual-view capabilities (a) and JPEG reconstruction fidelity at full gain (b). Metrics are averaged over 400 images per construction.}
\label{tab:reconstruction-quality}
\begin{tabular}{@{}lccc@{}}
\multicolumn{4}{@{}l}{\textbf{(a) Mechanism and representation capabilities}} \\
\toprule
Method & Mechanism & RGB pairing & Carrier \\
\midrule
AlphaDog & Alpha blend & \ding{55} & RGBA PNG \\
RGB & RGB gain & \ding{51} & JPEG / PNG \\
Single-channel & Shared gain & \ding{55} & JPEG / PNG \\
\bottomrule
\end{tabular}
\par\smallskip
\begin{tabular}{@{}lrrr@{}}
\multicolumn{4}{@{}l}{\textbf{(b) Reconstruction fidelity at full gain (JPEG)}} \\
\toprule
Construction & MAE $\downarrow$ & PSNR (dB) $\uparrow$ & Linear RMSE $\downarrow$ \\
\midrule
RGB & 1.27 & 43.03 & 0.0097 \\
Single-channel & 0.19 & 55.00 & 0.0023 \\
\bottomrule
\end{tabular}
\par\smallskip
\begin{minipage}{\columnwidth}
\footnotesize
\textit{Note:} PSNR $>40$\,dB generally indicates high quality~\cite{bull2021picturequality}.
\end{minipage}
\end{table}

\begin{figure}[t]
  \centering
  \includegraphics[width=\columnwidth]{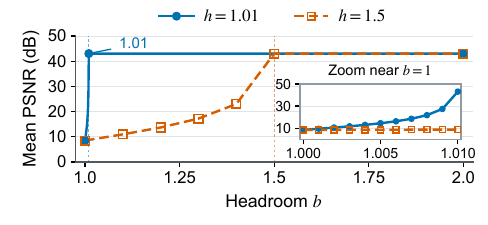}
  \caption{Mean reconstruction PSNR over the same 400 RGB images for $h=1.01$ and $h=1.5$. Both groups use $p=1$; only the full-gain threshold metadata differs. The inset enlarges $1\leq b\leq1.01$.}
  \label{fig:headroom-response}
\end{figure}

\paragraph{Fidelity across display headroom.}
The available display headroom varies across device hardware and software environments. 
Below the full-gain threshold $h$, the HDR image may retain visible SDR content. 
In practice, DRJanus sets the full-gain threshold $h$ to 1.01 to reduce the headroom required for complete HDR reconstruction. 
We compared HDR reconstruction quality for $h=1.01$ and $h=1.5$ at display headroom values below and above each threshold. 
We evaluated all 400 RGB images at selected headroom values within $b\in[1,2]$. 
Both configurations use $p=1$ and keep the SDR base image and gain-map encoding unchanged. 
Figure~\ref{fig:headroom-response} shows the average PSNR under each headroom setting.

As the available display headroom $b$ increases from 1 to 1.5, the average PSNR of the HDR reconstructions of DRJanus images with $h=1.5$ rises from 8.51 to 43.03 dB. 
At $b=1.2$, matching the default headroom measured on our Apple XDR display, the $h=1.01$ and $h=1.5$ configurations achieve average PSNR values of 43.03 and 13.70 dB, respectively. 
Both settings reach the same PSNR at full gain, but $h=1.5$ requires more display headroom. 
When headroom falls short, the gain is not fully applied and reconstruction quality can drop sharply. 
The supplementary material shows decoded views in which the target HDR content is mixed with residual SDR base content at intermediate headroom settings.

These results confirm that by selecting a lower full-gain threshold, DRJanus can reduce the headroom required for complete reconstruction and narrow the range in which the SDR base image may be exposed, while maintaining HDR reconstruction quality.

\paragraph{Human-perceived fidelity.}
We randomly sampled 20 original--reconstructed image pairs from the evaluation set and conducted a visual comparison with 10 participants using HDR-capable devices running macOS, Windows, Android, and iOS.
Each participant viewed DRJanus images generated with $h=1.01$ and $h=1.5$ alongside their corresponding original target images. 
For $h=1.01$, none reported visual anomalies or differences from the originals, and none recognized the images as HDR. 
All 10 participants reported abnormal visual transitions for $h=1.5$. 
These observations support the choice of $h=1.01$ to reduce perceptible transition artifacts and improve visual camouflage under the tested viewing conditions.

\subsection{RQ2: Moderation evasion}
\label{sec:moderation-understanding}

For moderation evasion, we pair a benign SDR base with harmful HDR content and evaluate whether moderation services and VLMs detect the harmful content.

\paragraph{Moderation services.}
Each service reports sexual content in a different form, so we converted each output to a binary flag as follows.
For OpenAI Moderation, we used the sexual-category flag.
For Amazon Rekognition, we flagged Explicit labels with confidence $\geq50\%$, following the API's default minimum confidence threshold.
For Azure AI Content Safety~\cite{azureImageContentSafety}, we flagged Sexual severity of Low or higher.
For Google Cloud Vision SafeSearch, we flagged an adult likelihood of \texttt{LIKELY} or \texttt{VERY\_LIKELY}.
For Tencent Cloud Image Moderation~\cite{tencentImageModeration}, we flagged a Porn recommendation of Review or Block.
For Alibaba Cloud Image Moderation~\cite{alibabaImageModeration}, we flagged medium or high pornographic risk.

As shown in Table~\ref{tab:moderation-main}, all six services detected every target in the HDR controls, while neither benign SDR base was flagged. None detected sexual content in the DRJanus images under either construction. 
Their decisions match the benign base, leaving the harmful HDR content undetected. By comparison, AlphaDog evaded five services but was detected by Tencent.

\begin{table}[t]
\centering
\footnotesize
\setlength{\tabcolsep}{2pt}
\caption{Sexual-content detection rates over 100 targets per condition. $\dagger$: API safety block, counted as detection. DRJanus: both RGB and single-channel gain-map JPEG constructions.}
\label{tab:moderation-main}
\begin{tabular}{@{}lccc@{}}
\toprule
System & \shortstack{HDR\\(\%)} & \shortstack{AlphaDog\\(\%)} & \bfseries\shortstack{DRJanus\\(\%)} \\
\midrule
\multicolumn{4}{l}{\textit{Moderation services}} \\
OpenAI~\cite{openaiModeration} & 100 & 0 & \textbf{0} \\
Amazon~\cite{amazonRekognitionModerationThreshold} & 100 & 0 & \textbf{0} \\
Azure~\cite{azureImageContentSafety} & 100 & 0 & \textbf{0} \\
Google~\cite{googleVisionSafeSearch} & 100 & 0 & \textbf{0} \\
Tencent~\cite{tencentImageModeration} & 100 & 100 & \textbf{0} \\
Alibaba~\cite{alibabaImageModeration} & 100 & 0 & \textbf{0} \\
\midrule
\multicolumn{4}{l}{\textit{Vision-language models}} \\
gpt-6-astra~\cite{openaiModels} & 100 & 100 & \textbf{0} \\
claude-opus-5-5~\cite{anthropicClaudeModels} & 100 & 100 & \textbf{0} \\
gemini-4-argon~\cite{googleGeminiModels} & 100 & 100 & \textbf{0} \\
qwen3.8-max-2026-09-02~\cite{alibabaQwenModels} & 100$^\dagger$ & 100 & \textbf{0} \\
doubao-seed-2-1-pro~\cite{volcengineDoubaoModels} & 100 & 100 & \textbf{0} \\
glm-5v-turbo~\cite{zaiGLM5VTurbo} & 100 & 100$^\dagger$ & \textbf{0} \\
kimi-k3~\cite{moonshotKimiK3} & 100 & 100 & \textbf{0} \\
deepseek-v4.1-flash-1017~\cite{deepseekModels} & 100 & 100 & \textbf{0} \\
grok-4.7~\cite{xaiGrokModels} & 100 & 0 & \textbf{0} \\
\bottomrule
\end{tabular}
\end{table}

\paragraph{Vision-language models.}
We evaluated nine VLMs on the same test set using a fixed content-safety classification prompt.
We count responses identifying explicit sexual content, suggestive content, or nudity as detections, and treat API safety blocks the same way.

As shown in Table~\ref{tab:moderation-main}, all nine models detected every target in the HDR controls but classified every DRJanus image as non-sensitive, consistent with the benign SDR base.
By comparison, eight models detected sexual content in the AlphaDog images; only \texttt{grok-4.7} classified them as non-sensitive.
Additional observations from browser interfaces and native clients are provided in the supplementary material.

\subsection{RQ3: Agent security in real-world scenarios}
\label{sec:agent-security}

\paragraph{Setup.}
We evaluate each agent in two threat settings, each with 120 cases (four cases in each of 30 real-world scenarios); scenario distributions and protocols are provided in the supplementary material.
In \emph{moderation evasion}, the user asks the agent to prepare a newsletter, update a product catalog, or complete a similar task using only images suitable for a general audience. 
We supply DRJanus images that show appropriate content in SDR but sexual content in HDR. 
We verify publication through the local service by checking that the designated image's original bytes and gain map are preserved.
In \emph{task manipulation}, the user asks the agent to send emails, schedule events, or select products based on information shown in images. 
DRJanus presents the intended task information in HDR but alters recipients, document requirements, or other details in the SDR base. 
An attack succeeds if the agent records an action or output that meets the attacker's objective and violates the user's requirements.
We do not evaluate AlphaDog in these settings, since eight of the nine VLMs in RQ2 already detected it.

\begin{table}[t]
\centering
\footnotesize
\setlength{\tabcolsep}{3pt}
\caption{Verified attack success rates over 120 cases per agent and threat setting.}
\label{tab:agent-outcomes}
\begin{tabular*}{\columnwidth}{@{\extracolsep{\fill}}lcc@{}}
\toprule
Agent (model) & \shortstack{Moderation\\evasion (\%)} & \shortstack{Task\\manipulation (\%)} \\
\midrule
Codex (\texttt{gpt-6.1-sol}) & 100.00 & 100.00 \\
\shortstack[l]{Claude Code\\(\texttt{claude-opus-5-5})} & 98.33 & 100.00 \\
ZCode (\texttt{glm-5.3-flash}) & 100.00 & 100.00 \\
Kimi Code (\texttt{kimi-k3}) & 100.00 & 100.00 \\
\shortstack[l]{DSH\\(\texttt{deepseek-v4.1-flash})} & 31.67 & 95.00 \\
\bottomrule
\end{tabular*}
\end{table}

\paragraph{Results.}
As shown in Table~\ref{tab:agent-outcomes}, Codex, ZCode, and Kimi Code reached 100\% attack success in both settings, and Claude Code reached 98.33\% and 100\%.
These agents acted on the SDR view. 
In both settings, Claude Code, ZCode, and Kimi Code rechecked image details, but these checks generally examined only the SDR view.
The exceptions were two moderation cases, in which unusually large files prompted Claude Code to inspect the gain map and detect the hidden content.

DSH was the most resistant agent, with attack success of 31.67\% in moderation evasion and 95\% in task manipulation, but its runs incurred higher overall workflow costs.
In moderation evasion, it repeatedly reassessed image suitability and inspected metadata, resisting 82 of the 120 attacks, and averaged 42 tool calls and 7.8 minutes per run across all 120 cases, versus 7.9--11.2 calls and 0.8--1.1 minutes for the other four agents.
In task manipulation, it recovered the task information in the HDR view and resisted six task-manipulation attacks; on these same six cases, it averaged 80.8 tool calls and 23.9 minutes per run, versus 5.8--7.7 calls and 0.4--1.7 minutes for those four agents.

DRJanus can also conceal inputs that are not meaningful to humans.
As a white-box demonstration beyond our threat model, we placed an adversarial image optimized against Qwen3.8-27B in the SDR base and a benign meme in the HDR view; the model produced the attacker's target sentence, whereas the HDR view elicited a reply about the meme (details in the supplementary material).

\section{Defense Against DRJanus}
\label{sec:defense}
A straightforward defense is to decode both views and check each independently for harmful content. Such checks can expose harmful HDR content, but cannot establish cross-view consistency: individually benign views may still convey different instructions, task information, or visual content that changes an agent's decisions. We therefore compare content across the SDR base and HDR reconstruction.

\paragraph{Cross-view consistency.}
We compare the SDR base $B$ with its full-gain HDR reconstruction $\widehat H_h$. Direct pixel comparison would treat legitimate HDR enhancement as a mismatch. We instead apply a normalized logarithmic transform to decoded linear RGB and subtract the spatially smoothed RGB mean. This suppresses shared luminance variation while retaining structural and chromatic changes. Light smoothing reduces encoding noise. Following structural image comparisons~\cite{wang2004ssim,yeganeh2013tonemapped}, we compute local disagreement from the processed features $X$ and $Y$:
\begin{equation}
\begin{aligned}
d(x)&=1-\operatorname{clip}_{[0,1]}
\frac{\sigma_{XY}+\eta}
{\sqrt{(\sigma_X^2+\eta)(\sigma_Y^2+\eta)}},\\
S&=\max_x d(x).
\end{aligned}
\label{eq:local-consistency}
\end{equation}
The variances and covariance use Gaussian-weighted neighborhoods and all RGB channels; $\eta$ stabilizes uniform regions. Taking the maximum prevents a small edit from being diluted by unchanged surroundings. We flag $S>\tau$, with $\tau\approx0.48424$ calibrated on 220 benign HDR scores.

\paragraph{Evaluation.}
The normal calibration data comprise 100 color and 100 grayscale HDR images generated from COCO~\cite{lin2014coco} using smooth highlight enhancement that preserves scene content, plus 20 public HDR samples~\cite{rahmanUltraHDRSamples,androidHDRSamples,libultrahdrTestImages}. To measure false alerts, we use 134 original camera HDR JPEGs from GSMArena and Notebookcheck review galleries (e.g.,~\cite{cameraOriginalsGsmarenaOneplus13,cameraOriginalsNbOneplus13}), spanning eight phone models. These files retain their original resolution, gain maps, and metadata.

We evaluate 600 DRJanus images from 200 COCO test sources disjoint from the COCO calibration sources, covering RGB whole-image replacements, local edits to email addresses or prices, and the single-channel construction. All attacks are detected, with one false alert among the camera HDR originals (Table~\ref{tab:defense-consistency}).

\begin{table}[t]
\centering
\small
\setlength{\tabcolsep}{5pt}
\caption{DRJanus attack detection and camera false alerts.}
\label{tab:defense-consistency}
\begin{tabular}{@{}lrr@{}}
\toprule
Images & Flagged / Total & Rate (\%) \\
\midrule
Whole-image replacements & 200/200 & 100.00 \\
Local edits & 200/200 & 100.00 \\
Single-channel attacks & 200/200 & 100.00 \\
\midrule
Camera HDR originals & 1/134 & 0.75 \\
\bottomrule
\end{tabular}
\end{table}

\paragraph{Adaptive attacks.}
We further test whether an attacker who knows the detector can evade it while preserving different text in the two views. Among 100 character-edit cases, only one evades detection with the text difference verified by OCR, indicating limited evasion in the tested settings.

\section{Discussion and Conclusion}

\paragraph{Limitations.}
DRJanus relies on gain maps being preserved during delivery and reconstructed by an HDR-capable viewer.
SDR-only viewers display the base image, and platforms that discard gain maps or convert uploads to SDR remove the dual-view effect.
However, as HDR support and usage expand across smartphones, browsers, and social platforms~\cite{appleAdaptiveHDR2024,webkitHDRImages2025,googleInstagramUltraHDR2024}, these conditions are increasingly met, and the attack surface is likely to grow.

\paragraph{Societal impacts and responsible use.}
The discrepancy between the SDR and HDR renditions undermines content safety and informed consent.
Harmful HDR content can bypass moderation, enabling unwanted exposure and targeted harassment, with particular risks for minors.
In agent workflows, users authorize actions based on one view while agents act on the other.
Services should therefore check both views for consistency, e.g., with our cross-view check (Section~\ref{sec:defense}), or provide users and agents with the same verified rendition.
All platform experiments used our own test accounts, with posts set to private or shared only among test accounts, so no harmful content reached other users.
We disclosed our findings to moderation services, social media platforms, smartphone vendors, and open-source library maintainers; a major social media platform rated the issue high severity and awarded a bug bounty.

\paragraph{Conclusion.}
We introduced DRJanus, which inverts gain-map reconstruction to encode independently chosen SDR and HDR images in one JPEG without target-model queries or optimization.
Experiments demonstrate high-fidelity reconstruction, moderation evasion across six services and nine VLMs, and verified harmful-image publication and task manipulation across five agents.
Our cross-view consistency check detects all 600 tested attacks, covering whole-image replacements, local edits, and single-channel constructions.
These results show why visual safeguards must account for both what users see and what systems act on: assessing one rendition cannot establish the safety or meaning of another.

\clearpage
\begingroup
\small
\raggedright
\bibliographystyle{unsrtnat}
\bibliography{main}
\endgroup

\end{document}